\documentclass[conference]{IEEEtran}

\IEEEoverridecommandlockouts
\usepackage{cite}
\usepackage{amsmath,amssymb,amsfonts}
\usepackage{algorithmic}
\usepackage{textcomp}
\usepackage{xcolor}
\usepackage{subcaption}  
\usepackage{graphicx}

\usepackage{caption}
\usepackage{varwidth}

\usepackage[utf8]{inputenc} 
\usepackage[T1]{fontenc}    
\usepackage{url}            
\usepackage{booktabs}       
\usepackage{amsfonts}       
\usepackage{nicefrac}       
\usepackage{microtype}      
\usepackage{xcolor}         

\usepackage{graphicx}
\usepackage{xcolor}
\usepackage{amssymb}

\usepackage{rotating}

\newlength{\fheight}
\newcommand{\outlineN}[1]{}
\newcommand{\outline}[1]{}
\newcommand{\format}[1]{{\fontfamily{qcr}\selectfont #1}}
\usepackage{soul}

\usepackage{caption}
\usepackage{subcaption}
\usepackage{varwidth}
\DeclareCaptionFormat{myformat}{%
  \begin{varwidth}{\linewidth}%
    \centering
    #1#2#3%
  \end{varwidth}%
}

\usepackage{enumitem}
 
 \usepackage{multicol}
 
\usepackage{titlesec}

\usepackage{siunitx}
\usepackage{todonotes}
\usepackage{gensymb}

 \usepackage{fancyhdr} 
 \usepackage{setspace} 

\def\BibTeX{{\rm B\kern-.05em{\sc i\kern-.025em b}\kern-.08em
    T\kern-.1667em\lower.7ex\hbox{E}\kern-.125emX}}

\begin{document}

\title{A Medical Low-Back Pain Physical Rehabilitation Dataset for Human Body Movement Analysis 
\thanks{This work  is partially supported by the EU FP7 grant ECHORD++ KERAAL and by the European Regional Fund via the VITAAL Contrat Plan Etat Region.}
}

\author{
          \IEEEauthorblockN{Sao Mai Nguyen}
               \IEEEauthorblockA{IMT Atlantique\\ Lab-STICC, UMR 6285\\
              and  FLOWERS U2IS, \\ ENSTA, IP Paris \& Inria,\\ France\\
              {nguyensmai@gmail.com}   }        
            \and
            \IEEEauthorblockN{Maxime Devanne}
              \IEEEauthorblockA{Université de \\Haute-Alsace \\
              IRIMAS EA 7499, \\
              France\\
              }          
          \and

            \IEEEauthorblockN{Olivier Remy Neris} 
            \IEEEauthorblockA{Université Brest\\ CHU Brest\\ INSERM, UMR 1101\\ F 29200 Brest, \\France
            }
          \and

            \IEEEauthorblockN{Mathieu Lempereur} 
            \IEEEauthorblockA{Université Brest\\ CHU Brest\\ INSERM, UMR 1101\\ F 29200 Brest, \\France
            }
            \and
              \IEEEauthorblockN{Andr\'e Th\'epaut}
              \IEEEauthorblockA{IMT Atlantique\\ Lab-STICC, UMR 6285,\\
              F-29238 Brest,\\
              France}
}


\maketitle
\thispagestyle{fancy}
 \lhead{}
 \chead{
 \texttt{
 \scriptsize{
 Nguyen, S. M., Devanne, M., Remy-Neris, O., Lempereur, M., and Thepaut, A. (2024). A Medical Low-Back Pain Physical Rehabilitation Dataset for Human Body Movement Analysis. International Joint Conference on Neural Networks. 
  }
 }
 \vspace{20pt}}
 \rhead{}

\begin{abstract}
  While automatic monitoring and coaching of exercises are showing encouraging results in non-medical applications, they still have limitations such as errors and limited use contexts.  To allow the development and assessment of physical rehabilitation by an intelligent tutoring system, we identify in this article four challenges to address and propose a medical dataset of clinical patients carrying out low back-pain rehabilitation exercises. The dataset includes 3D Kinect skeleton positions and orientations, RGB videos, 2D skeleton data, and medical annotations to assess the correctness, and error classification and localisation of body part and timespan. Along this dataset, we perform a complete research path, from data collection to processing, and finally a small benchmark. We evaluated on the dataset two baseline movement recognition algorithms, pertaining to two different approaches: the probabilistic approach with a Gaussian Mixture Model (GMM), and the deep learning approach with a Long-Short Term Memory (LSTM).
  This dataset is valuable because it includes rehabilitation relevant motions in a clinical setting with patients in their rehabilitation program, using a cost-effective, portable, and convenient sensor, and because it shows the potential for improvement on these challenges.
\end{abstract}


\section{Introduction}
\label{intro}

\begin{figure}[!b]
\centering
\vspace{-0.3cm}
\begin{subfigure}{.15\textwidth}
\centering
  \includegraphics[height=0.8\fheight]{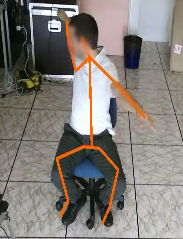}
\vspace{-0.2cm}
  \caption{torso rotation}
  \label{}
\end{subfigure}
\hfill
\begin{subfigure}{.15\textwidth}
\centering
  \includegraphics[height=0.8\fheight]{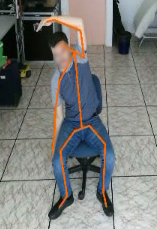}
\vspace{-0.2cm}
  \caption{flank stretch}
  \label{}
\end{subfigure}
\hfill
\begin{subfigure}{.15\textwidth}
\centering
  \includegraphics[height=0.8\fheight]{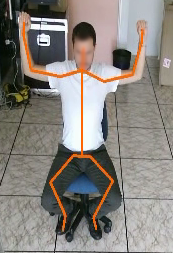}
\vspace{-0.2cm}
  \caption{hiding face}
  \label{}
\end{subfigure}
\vspace{-0.2cm}
\caption{The three rehabilitation exercises in our dataset}
\label{fig:exercises}
\end{figure}

\outlineN{ 1p. We define the new challenges. We present a dataset and an initial benchmark. In addition to provifing a performance benchmark for multi-agent behavior classification, our dataset is suitable for studying several research questions, including : ...}

\paragraph{Physical rehabilitation} 
Back pain is the 3rd common musculoskeletal disabling condition in the 45-65 years old population, and therapists consider that regular physical exercises is essential to alleviate back pain. 50 to 80\% of the world population suffers at a given moment from back pain \cite{Mounce2002R,WHO2003}, caused by  accident, surgery, old age or unfit working habits. 
While active rehabilitation is considered more effective for physical rehabilitation than usual care \cite{Kent2012T,Everard2021NNR}, therapists are concerned about the decreasing engagement of the patients throughout the months-long repetition of physical exercises \cite{Gordon2016H},  primarily attributed to the lack of supervision and monitoring of the patient performance.

 To promote physical activity, games \cite{Pasco2011SSR}, mobile phone applications \cite{Kermarrec2015EPSCIKDSEHSSR}, virtual agents \cite{Waltemate2015PSVRSTV,Anderson2013ACE} and robots \cite{Fasola2013JHI,Devanne2018PIICRC,Hun-Lee2023UMUI,Feingold-Polak2021JNR} can enhance the engagement of the patient. These approaches  rely on body motion analysis of patients. 
 Leveraging machine learning and computer vision, 3D physical body movements analysis has improved the performance of its algorithms. 
 Unfortunately, their performance relies on the quality of the training dataset, and most human motion datasets have not been carried out in the medical context \cite{Liao2020CBM}:  they lack medical annotation and variability representative of patients, and most contain data from only healthy subjects.

\paragraph{Challenges}
Our main motivation for creating the presented dataset is to create an intelligent tutoring system capable of supervising rehabilitation sessions autonomously such as presented in \cite{Blanchard2022BRI}, by providing instructions for each exercise of the program and real-time feedback how to improve the efficiency of the patient's performance. 
With the aim of rehabilitation using human body movement analysis, intelligent tutoring systems (ITS) need to analyse complex full-body exercises that can involve several parts of the body but not necessarily all parts of the body. The assessment algorithm should be able to understand which parts are important, and what are the ranges of freedom that are acceptable. The ITS should  encapsulate the tolerated variance for each joint and time frame. Thus, we have identified 4 challenges in rehabilitation movements analysis:

 \begin{enumerate}
     \item Rehabilitation motion assessment. The goal is to assess an observed motion sequence by detecting if the rehabilitation exercise is correctly performed or not.
     \item Rehabilitation error recognition. The goal is to classify the observed error among a set of known errors, so as to explain and give feedback. 
     \item Spatial localization of the error. In addition to recognizing the error, the goal is also to identify which body part is responsible of the error. 
     \item Temporal localization of the error. The goal is to detect the temporal segment where the detected error occurred along the sequence.
 \end{enumerate}

 While most rehabilitation datasets only provide annotations for the rehabilitation motion assessment, we propose a medical dataset of clinical patients carrying out low back-pain rehabilitation exercises. The dataset  includes 3D Kinect skeleton positions and orientations, RGB videos, 2D skeleton data, and medical annotations to assess the correctness, label and timing errors of each movement part. The article also provides initial benchmarks with 2 movement analysis algorithms.
 
\section{Related Work}
\label{sec:soa}

\paragraph{Capture system} 
A number of human activity datasets have been recently reported. They were mostly captured in two modalities: marker-based motion capture (MoCap) tracking and vision-based (marker-less) tracking \cite{Liao2020CBM}. On the other hand, vision-based technologies 
 do not hinder movements. RGB-D cameras, such as the Microsoft Kinect, are affordable, require little calibration or setup. Moreover, the Kinect has been validated against standard motion capture systems despite a lower precision, and against other vision-based tracking algorithms, and wearable sensors \cite{Clark2012GP,Yang2014ISJ,Stone2011JAISE,Jebeli2017NE,Faity2022S,Xu2017AE}. 
More recently, human pose estimation from RGB images \cite{Cao2019ITPAMI,Xiu2018,Sun2019PICCVPR,Bazarevsky2020C,Guler2018PICCVPR,Zhen2020CVTE2,Benzine20202ICCVPRC} have also made impressive improvements
, leveraging deep learning models and large datasets. 
Cameras can easily blend into patient's homes or a clinical environment.
They are affordable, require little calibration or setup. This makes them suitable for use in our clinical test where patients are recorded in daily sessions of 30 minutes.
Thus, we provide data from both RGB cameras and the Kinect.

\begin{table*}[tb]
\caption{Comparison with other datasets (Participants column reports the total number of participants including the patients. NA = Not Available, annotators column report the labelling method : if the cases were recorded under instructions or if a medical or non medical human annotator labelled the data )}
\label{tab:comparison}       
\begin{tabular}{p{1cm} p{2cm} p{1.7cm} p{0.5cm} p{0.5cm} p{0.5cm} p{0.5cm} p{1.2cm} p{0.8cm} p{0.5cm} p{3.2cm} p{1.5cm}}
\hline
Dataset  & Activities & Exercises & Nb exer. & Pat-ients & Par-tici-pant\footnote{including patients} & EMG & RGB D JP JO A\footnote{RGB videos/images, Depth, Joint Position, Joint Orientation, Audio}& MoCap /wearable & Nb record. & Annotations & Annotators \\
\hline\noalign{\smallskip}
\\ 

K3Da ~\cite{Leightley20152ASIPAASCA} & clinical assess. of gait, balance & posture tests & 13 & 0 & 54 & NA & Kinect2 D, JP & NA & 525 & None & None \\
EMG Squat \cite{Nishiwaki2006JJPTA} & Therapy to protect the anterior cruciate ligament (ACL) & Squat & 3 & 0 & 9 & leg & NA & NA & 81 & Exercise type & instructions \\

HPTE \cite{Ar2014ITNSRE} & Physiotherapy exercises at home & Shoulder and knee exercises & 8 & 0 & 5 & NA & Kinect1 RGB D& NA & 240 & Exercise type & instructions \\

%
UI-PRMD \cite{Vakanski2018D} &  Physical therapy and rehabilitation & Whole-body exercises & 10 & 0 & 10 & NA & Kinect1JP JO& Vicon & 1000 & Correct/ Incorrect &
instructions\\

TRSp \cite{Dolatabadi2017P1ICPCTH} & Post-stroke physical rehab  & Arm movements using a haptic robot tabletop & 2 & 9 & 19 & NA & Kinect1 JP JO & NA & 190 & Error label & 2 Med \\

IRDS \cite{Miron2021D}& general physical rehabilitation &  
arms and legs &
9 & 15 & 29&
NA &
Kinect2 JP &
NA &
2589 &
Exercise type position (sitting or standing), correctness label (correct, incorrect, unrecognisable) &
2 Non Med\\

EmoPain \cite{Aung2016ITAC} & Chronic low back pain & Physical exercises & 7 & 22& 50 & back  & multi-RGB A & IMU  & unk. & pain expression \& pain related mvt&
8 Non Med + 6 Med 
\\

KIMORE \cite{Capecci2019ITNSRE} & low back pain & 
whole body &
5 & 34 & 78 & NA & 
Kinect2 RDB D JP JO & NA & 1000 &
3 scores (quantify the accuracy of the subjects) for total error and body part & 
3 Med\\

Keraal & Posture and back pain rehabilitation & Upper-body exercises & 3 & 12 & 21 & NA & Kinect2 RGB, JP, JP & Vicon (Gr3) &  2622 & 
Error label + (Group1a,2a: body part, timespan)&
2 Med \\
 
\noalign{\smallskip}\hline
\end{tabular}
\vspace{-0.2cm}
\end{table*}

\paragraph{Medical Human Movement Dataset}
Although several datasets contribute to research in human motion analysis, such as the Kinect datasets \cite{Wu2015IICCVPR,Koppula:2013b,Shahroudy2016IICCVPR,Liu2017APA} 
or the multi-sensor datasets \cite{Stein2013PIWMCEA,Tenorth2009IIWTHETMISTCWI,Roggen20102SICNSSI}, 
they can hardly be applied in the medical context, including physical rehabilitation.

The K3Da dataset~\cite{Leightley20152ASIPAASCA} 
  is the first Kinect based dataset in a healthcare setting based on common clinical assessments of gait and balance. Nevertheless, they did not specifically recruit patients and the data were not labelled with medical criteria. 
The HPTE 
dataset  \cite{Ar2014ITNSRE} records with a Kinect  therapy movements for computerized monitoring at home of 8 shoulder and knee exercise movements. 
Again, the participants are  healthy. 
 The EMG Squat dataset  \cite{Nishiwaki2006JJPTA} is restricted to EMG electrodes recordings of 3 exercises of lower limbs performed by 9 healthy participants.
   The 
  UI-PRMD \cite{Vakanski2018D} proposes a dataset of 
  10 healthy subjects performing 
  physical therapy recorded by a Vicon marker-based tracker and a Kinect
  . Moreover, the labeling only indicates if the execution is correct, but not the type of error or their timing.
  
  \paragraph{Human Movement Dataset with Patients} 
Targeting low hack pain too, the EmoPain dataset \cite{Aung2016ITAC}, to study the effects of pain during physical rehabilitation, provides high-resolution face videos, audios, full body joint motions, and electromyographs (EMG) from back muscles. 
   The labels include pain facial expressions and pain-related movements. Like K3Da, EmoPain is labeled for recognition of exercises or mental states, they do not address the challenges  stated in Sec. \ref{intro}.   
   The 
   TRSP dataset \cite{Dolatabadi2017P1ICPCTH} proposes a dataset of clinically relevant motions during robotic rehabilitation exercises focusing on strength exercises with a haptic robot tabletop, captured with a Kinect. Likewise, IRDS \cite{Miron2021D} provides Kinect data of patients doing general rehabilitation. In TRSP and IRDS, The data from both healthy and patients are assessed by clinicians with scores and error labels, so they can address the challenges 2 and 3.  But the labels do detail which body part and timespan of the execution was wrong. The Kimore dataset  \cite{Capecci2019ITNSRE} proposes 5 whole body exercises from patients and healthy subjects. 
  It is the closest to ours : it also targets low-back pain and is labelled by medical annotators. The labels are both at the global and the body part level  to address challenges 1, 2 and 3. However, Kimore does not yet provide with the temporal information of errors to address challenge 4.

In comparison with these datasets, our Keraal dataset has been recorded within a long-term rehabilitation program, targeting low back pain. 
 Contrarily to EMG Squat, HPTE and K3Da, UI-PRMD, we have recruited rehabilitation patients and have data labelled by a doctor. Like the HPTE dataset \cite{Ar2014ITNSRE}, our work is in a long-term effort of enabling physiotherapy exercises at home and our data is extracted from a 4-weeks evolution of each patient.  
A full comparison can be read in Table \ref{tab:comparison}. 
Thus our Keraal dataset is the only benchmarking set with clinical patients and with labels provided by a physician for the four challenges : motion assessment, error recognition, spatial  and temporal localization.


\section{Dataset and Framework}
\outlineN{Data collection,  Data structure and documentation.}

This section describes the protocol and rationale for how the Keraal dataset was created, the participants included, the hardware setup and the experimental protocol. The dataset and the code are available on \url{http://nguyensmai.free.fr/KeraalDataset.html} and \url{https://github.com/nguyensmai/KeraalDataset}.

\paragraph{Rehabilitation program}
31 patients, aged 18 to 70 years, were recruited in the double blind study. This prospective, centrally randomized, controlled, single-blind and bi-centric study was conducted from October 2017 to May 2019.
  12 patients suffering from low-back pain were included 
  and were asked 
   to perform each of the three predefined exercises the best they can from its demonstration. The details on this clinical trial, including the patient care, the rehabilitation sessions, 
   the inclusion and exclusion criteria, the characteristics of the patients, the efficiency of the care have been reported in \cite{Blanchard2022BRI}.  This study has received a Legal authorisation for clinical tests from the medical ethics board of the hospital at Brest (CHRU Brest). All subjects have given their informed consent to participate in the study.

\subsection{Exercises and errors}
\label{sec:listErrors}

A list of three exercises have been chosen in conjunction with therapists as common rehabilitation exercises that are also used for low-back pain treatment, under the condition that they can be coached by an intelligent tutoring system using visual assessment. Illustrations of these exercises can be seen in Fig. \ref{fig:exercises} 
. The 3 exercises are centered on spine stretching: a left rotation of the trunk followed by a the same right rotation, a left and right lateral bending of the trunk and a breathing exercise with the upper limbs flexed 90\degree at shoulder and elbow.

\begin{figure*}[t!]
\centering
\begin{subfigure}{.33\textwidth}
\centering
  \includegraphics[height=0.6\fheight]{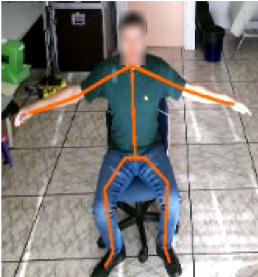}
  \caption{torso rotation -- error1:\\ Arms are not raised enough}
  \label{}
\end{subfigure}
\hfill
\begin{subfigure}{.3\textwidth}
\centering
  \includegraphics[height=0.6\fheight]{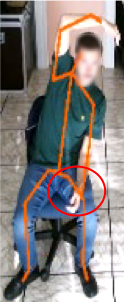}
  \caption{flank stretch -- error1: \newline Opposite arm is not along the body}
  \label{}
\end{subfigure}
\hfill
\begin{subfigure}{.3\textwidth}
\centering
  \includegraphics[height=0.6\fheight]{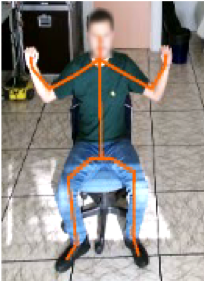}
  \caption{hiding face -- error1:\\ Arms are not raised enough }
  \label{}
\end{subfigure}
\\

\begin{subfigure}{.33\textwidth}
\centering
  \includegraphics[height=0.6\fheight]{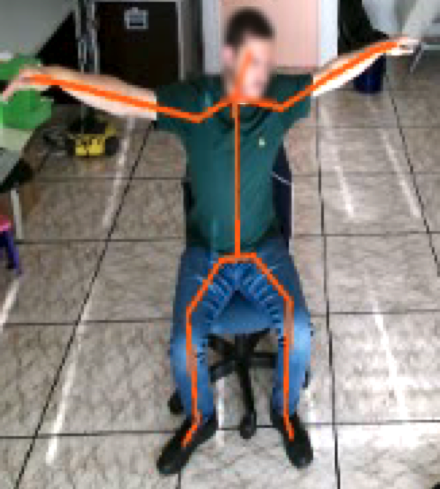}
  \caption{torso rotation -- error2: \\The torso's rotation is not sufficient}
  \label{}
\end{subfigure}
\hfill
\begin{subfigure}{.3\textwidth}
\centering
  \includegraphics[height=0.6\fheight]{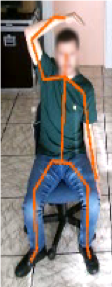}
  \caption{flank stretch -- error2:\\ Body is not tilted }
  \label{}
\end{subfigure}
\hfill
\begin{subfigure}{.3\textwidth}
\centering
  \includegraphics[height=0.6\fheight]{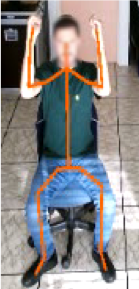}
  \caption{hiding face -- error2: \\Arms are not outspread enough }
  \label{}
\end{subfigure}
\\

\begin{subfigure}{.33\textwidth}
\centering
  \includegraphics[height=0.6\fheight]{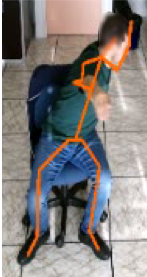}
  \caption{torso rotation -- error3: \\The body is leaned on the side }
  \label{}
\end{subfigure}
\hfill
\begin{subfigure}{.3\textwidth}
\centering
  \includegraphics[height=0.6\fheight]{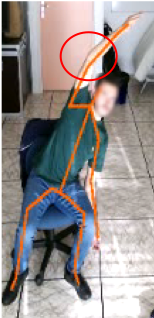}
  \caption{flank stretch -- error3:\\ The above arm is not bent }
  \label{}
\end{subfigure}
\hfill
\begin{subfigure}{.3\textwidth}
\centering
  \includegraphics[height=0.6\fheight]{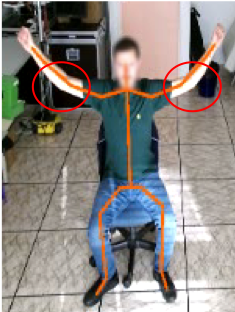}
  \caption{hiding face -- error3: \\Arms are not raised enough}
  \label{}
\end{subfigure}
\\
\caption{Each column illustrates 3 errors for each exercise.}
\label{fig:errEx}
\end{figure*}

A list of common errors was defined 
 in conjunction with the experience of the therapists of CHRU Brest. Errors are 
illustrated in Fig. \ref{fig:errEx}.


\captionsetup{format=myformat}

\subsection{Participants}\label{participants}
The dataset contains data from three groups of participants:
\begin{itemize}
\item{Group1 : Rehabilitation Patients} :
 
The data of the daily sessions of the 12 patients recruited 
 is split into two subsets:
\begin{itemize}
    \item Group1a: 14 recordings per exercise among 6 patients were annotated as detailed in Sec.\ref{sec:annotation}.  
    \item Group1b: the remaining recordings of the 12 patients without annotation.
\end{itemize}

\item{Group2 : Healthy participants with Kinect V2 recordings} : 
Six healthy adults 
performed the exercises after the same instructions. 
 They are free to execute the exercise correctly or with errors. 
As for group1, the recordings of the 6 healthy adults are split into two subsets:
\begin{itemize}
    \item Group2a: 51 recordings per exercise were annotated as detailed in Sec. \ref{sec:annotation}.  
    \item Group2b: the remaining recordings of the 6 healthy adults without annotation.
\end{itemize}

\item{Group3 : Healthy participants : }
Three healthy adults 
 performed correct execution of exercises and simulated the identified common errors described in Sec. \ref{sec:annotation}. 
\end{itemize}

\subsection{Sensor system}
Using the Microsoft Kinect V2 sensor, we obtained the RGB video with the skeleton drawn, and the skeleton joint positions and orientations information. From the RGB videos, we can also obtain additional estimation of joint positions and orientations using the human body keypoint detection libraries OpenPose \cite{Cao2017C} and Blazepose \cite{Bazarevsky2020C}.
Moreover, {as the Vicon system is considered the best system for precision,} for Group3, we also recorded with MoCap using the Vicon system. For synchronisation purpose, the two systems were activated simultaneously. 

Comparing the human pose estimation methods, \cite{Marusic2023C2AICHI} showed that OpenPose and BlazePose data lead to comparable performances of GMM to classify correct/incorrect exercises (challenge 1). Thus in the following, we report results with Kinect data.

\subsection{Dataset annotation}
\label{sec:annotation}
The videos collected from patients with the Kinect V2 skeleton drawn were annotated by a medical doctor in physiotherapy and a physiotherapist
, using the Anvil video annotation research tool\footnote{http://www.anvil-software.org}.
The videos without blurred faces) are annotated at three levels related to the 4 challenges described in Sec. \ref{intro}. On a global evaluation level, an assessment is given as either correct or incorrect (Challenge 1). In the case of an incorrect error, they can indicate if the execution has no errors but finished before the end (label code $4$: incomplete) or the participant did not start the execution of the exercise (label code $5$ : motionless). On the error classification level, in the case of an incorrect movement, annotations first indicate whether the error is significant or small as well as the label of the error (Fig. \ref{fig:errEx}) (Challenge 2). Moreover the body part causing the error is also indicated (Challenge 3). 
 On a temporal level, the annotators can also indicate the time window where the error occurs (Challenge 4), and the same information as previously: whether the error is significant or small, the  error label, and the body part causing the error. The annotation is carried out a frame level. 

For challenge 1 to annotate a movement as correct or incorrect, we carried out an interannotator agreement analysis. The results show substantial agreement between the two medical annotators ( Cohen's $\kappa$ = 0.63 and Krippendorff's $\alpha$ = 0.62 \cite{Krippendorff2018},\cite{Cohen1960EPM}). 

\subsection{Dataset description}
Our dataset is composed of :

\begin{itemize}
  
    \item The therapist's annotations 
     indicate  whether the execution is correct, the label of the error, the body-part causing the error and the temporal description of the beginning and ending timestamps of the error.
    \item Anonymised RGB videos. 
    \item The positions and orientations of each joint of the Microsoft Kinect V2 skeleton.   
    \item The 2D positions of each joint of the OpenPose, Alphapose and Blazepose skeletons in the COCO pose or the 33 3D landmarks output format
    \footnote{\url{https://github.com/CMU-Perceptual-Computing-Lab/openpose/blob/master/doc/02_output.md}}. 
    \item the positions and orientations of each joint of the Vicon skeleton. 
    
\end{itemize}

 We have summarized the data and the number of recordings available per participant group in Table \ref{tab:dataset}.

\begin{table*}
\vspace{-1em}
\caption{Available recording modalities per group and the format of the data}
\vspace{-0.5em}
\label{tab:dataset}       
{\small
\begin{tabular}{lp{5cm}lllll}
\hline
Gr. & Annotation & RGB videos & Kinect  & Openpose/Blazepose  & Vicon & Nb rec\\
\hline\noalign{\smallskip}
1a & \format{xml anvil} : err label, bodypart, timespan & \format{mp4}, 480x360 & \format{tabular} & \format{dictionary} & NA & 249\\ 
1b & NA &  \format{mp4}, 480x360 & \format{tabular} & \format{dictionary} & NA & 1631\\ 
2a & \format{xml anvil} : err label, bodypart, timespan & \format{mp4}, 480x360 & \format{tabular} & \format{dictionary} & NA & 51\\ 
2b & NA & \format{mp4}, 480x360 & \format{tabular} & \format{dictionary} & NA & 151\\ 
3 & error label &  \format{avi}, 960x544 & \format{tabular} & \format{dictionary} & \format{tabular}& 540\\ 
\noalign{\smallskip}\hline
\vspace{-1em}
\end{tabular}
\vspace{-1em}
}
\end{table*} 

The Keraal dataset has a total of 2622 recordings, with 1881 recordings of patients and 741 recordings of healthy participants. Therefore, our number of patients recording already outnumbers the total number of recordings of all the other datasets, but the IRDS dataset.


\section{Benchmarks of the Keraal Dataset}
\subsection{Two algorithms}
\begin{figure*}[h!]
\begin{subfigure}{.50\textwidth}
\centering
    \includegraphics[height=0.7\fheight]{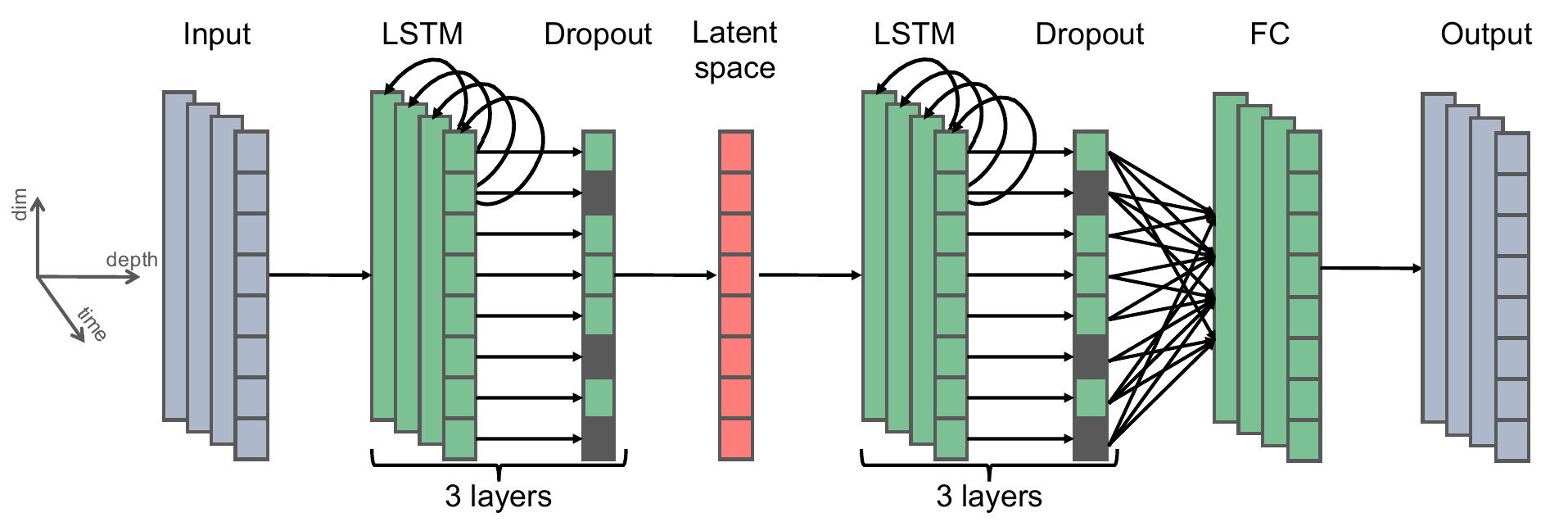}
    \caption{Architecture of the LSTM-autoencoder}
    \label{fig:lstm_ae}
\end{subfigure}
\hfill
\begin{subfigure}{.45\textwidth}
    \centering
    \includegraphics[height=0.7\fheight]{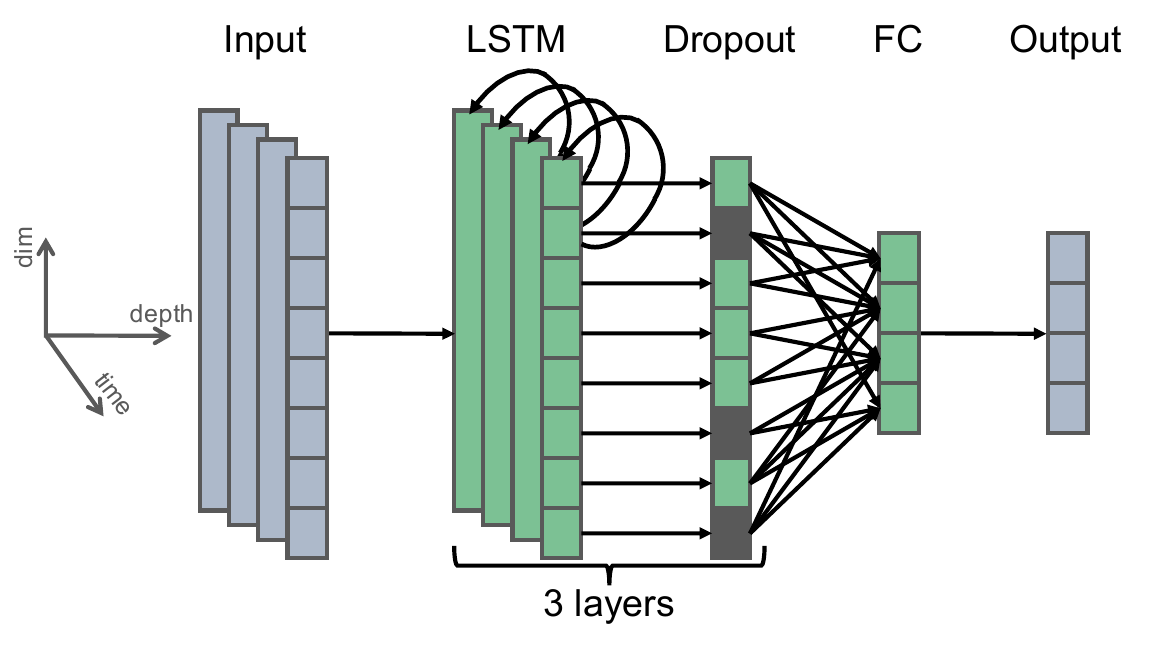}
    \caption{Architecture of the LSTM classifier}
    \label{fig:lstm_classif}
\end{subfigure}
\vspace{-0.5em}
\caption{LSTM models for assessment and classification}
\vspace{-1em}
\end{figure*}

In order to evaluate the dataset, we propose to assess the performance of human motion analysis algorithms in the context of rehabilitation. 
These baseline performances can serve as reference for future comparison with more sophisticated approaches. 
The first approach is a probabilistic approach uses a Gaussian Mixture Model (GMM) on a Riemannian Manifold~\cite{jost2005riemannian,zeestraten2017approach} 
. The second method is a deep learning-based approach employing a LSTM~\cite{hochreiter1997long} as applied for motion analysis in \cite{li2018independently,liu2017global} as the main part of its architecture.  We use a 3-layers LSTM with \textit{tanh} activation followed by dropout layer as the core of two different architectures employed for rehabilitation motion assessment and rehabilitation motion recognition. For both configurations, we use the Adam optimizer with a learning rate of $0.01$ and a batch size of $32$ on a simple laptop.  
These two approaches are evaluated and compared for the challenges of rehabilitation motion assessment (Challenge 1) and rehabilitation motion recognition (Challenge 2). 
 Position and orientation features of the upper body of the Kinect skeleton data are used in the following. 
For rehabilitation motion assessment, we employ our LSTM model within an autoencoder architecture in order to assess movements as correct or incorrect (Fig.~\ref{fig:lstm_ae}). For rehabilitation motion error classification, we add a fully-connected layer with \textit{softmax} activation 
 to train over $1000$ epochs (Fig. \ref{fig:lstm_classif}).

\subsection{Rehabilitation motion assessment}
\label{sec:evaluation}
The training data correspond to healthy subjects' correct demonstrations (Group3 cf. \ref{participants}), while testing data are participants' performances acquired during rehabilitation sessions. 
 and labelled (Group 1a and  2a cf. \ref{participants}). 
The unlabelled data from patients (Group 1b) can be used for unsupervised or semi-supervised learning.

\begin{figure*}[h!]
\begin{subfigure}[t]{.5\textwidth}
\begin{subfigure}[t]{.4\textwidth}
  \includegraphics[height=0.75\fheight]{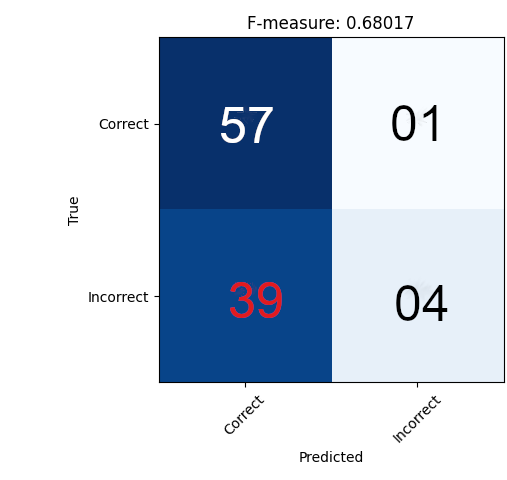}
  \caption{GMM torso rotation}
  \label{fig:sub-tsne-time}
\end{subfigure}
\hfill
\begin{subfigure}[t]{.27\textwidth}
  \includegraphics[height=0.75\fheight]{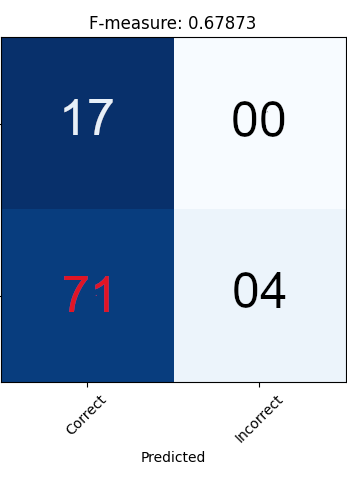}
  \caption{GMM flank stretch}
  \label{fig:sub-tsne-basic}
\end{subfigure}
\hfill
\begin{subfigure}[t]{.28\textwidth}
  \includegraphics[height=0.75\fheight]{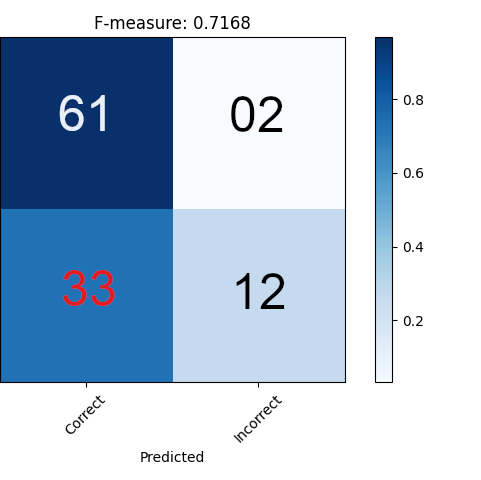}
  \caption{GMM hiding face}
  \label{fig:sub-tsne-mt}
\end{subfigure}
\end{subfigure}
\hfill
\begin{subfigure}[t]{.5\textwidth}

\begin{subfigure}[t]{.4\textwidth}
  \includegraphics[height=0.75\fheight]{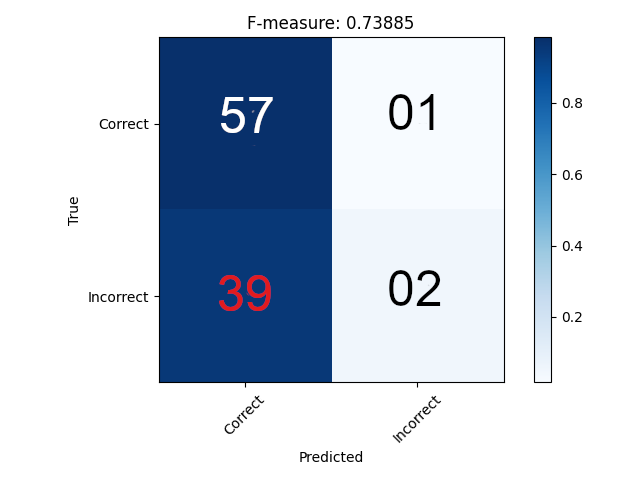}
  \caption{LSTM torso rotation}
  \label{fig:sub-tsne-time}
\end{subfigure}
\hfill
\begin{subfigure}[t]{.27\textwidth}
  \includegraphics[height=0.75\fheight]{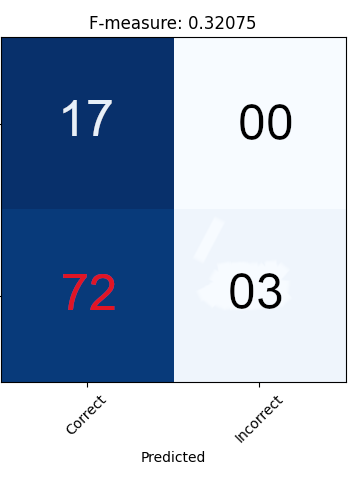}
  \caption{LSTM flank stretch}
  \label{fig:sub-tsne-basic}
\end{subfigure}
\hfill
\begin{subfigure}[t]{.28\textwidth}
  \includegraphics[height=0.75\fheight]{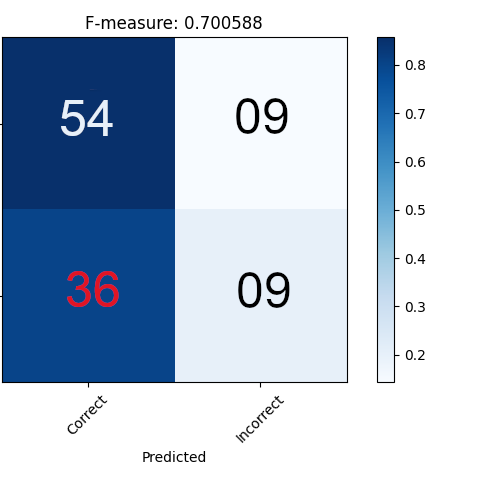}
  \caption{LSTM hiding face}
  \label{fig:sub-tsne-mt}
\end{subfigure}
\end{subfigure}
\vspace{-0.5em}
\caption{Confusion matrix of error detection using SVM and LSTM autoencoder.  }
\vspace{-1em}
\label{fig:confmat-gmm}
\label{fig:confmat-lstm}
\end{figure*}

We compare the ability of the two baselines to detect incorrect motion sequences while only correct demonstrations are available during training. 
 After determining by grid search the best threshold values for classification, Fig. \ref{fig:confmat-lstm} shows the detection results for the best F1-score of the GMM baseline and LSTM baseline, respectively. While colors represent normalized values, we left absolute values within the matrices so as to emphasize that classes are imbalanced.

\begin{figure*}

\begin{subfigure}{.4\columnwidth}
  \includegraphics[width=\columnwidth]{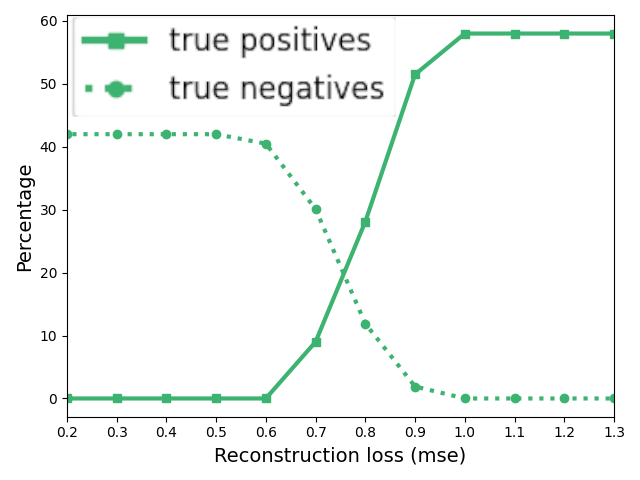}
  \caption{LSTM torso rotation}
  \label{fig:sub-tsne-time}
\end{subfigure}
\begin{subfigure}{.4\columnwidth}
  \includegraphics[width=\columnwidth]{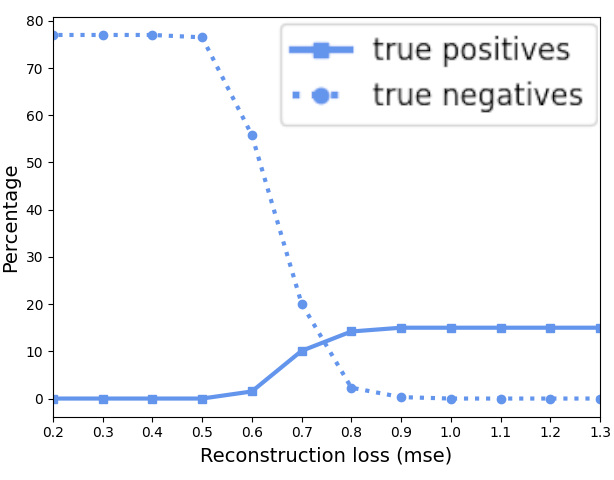}
  \caption{LSTM flank stretch}
  \label{fig:sub-tsne-basic}
\end{subfigure}
\begin{subfigure}{.4\columnwidth}
  \includegraphics[width=\columnwidth]{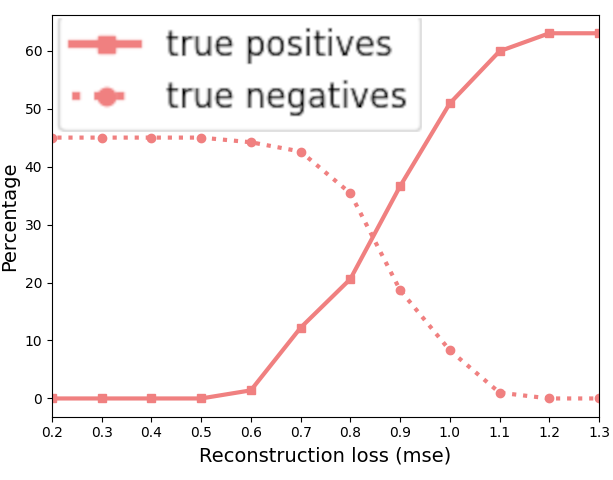}
  \caption{LSTM hiding face}
  \label{fig:sub-tsne-mt}
\end{subfigure}
\vspace{-0.5em}
\caption{LSTM baseline : true positives and true negatives evolution with respect to different thresholds employed to evaluate a movement as correct or incorrect.}
\vspace{-0.5em}
\label{fig:tptnLSTM}
 \label{fig:tptnGMM}
 \end{figure*}

Most of the sequences are classified as correct, as shown Fig. \ref{fig:confmat-gmm}. The majority of incorrect motion sequences are misclassified as correct, as highlighted in red color. This is a critical point showing that the proposed baselines may not be appropriate for rehabilitation motion analysis as it does not allow patients to improve their performance. This suggests that selecting a lower threshold may allow the approaches to correctly detect errors with the cost of also misclassifying correct sequences as incorrect. Depending on the requirement, a trade-off can be chosen. To further facilitate the choice, we show in Fig. \ref{fig:tptnGMM} 
 the evolution of the number of true positives (correct motion) and true negatives (incorrect motion) with respect to the thresholds for both baselines.

Further, \cite{Marusic2023ECMR} studied the impact of the amount of training data on this performance for GMM and Spatio-Temporal Graph Convolutional Networks with this dataset.

\begin{figure*}
\begin{subfigure}{.4\columnwidth}
  \includegraphics[width=\columnwidth]{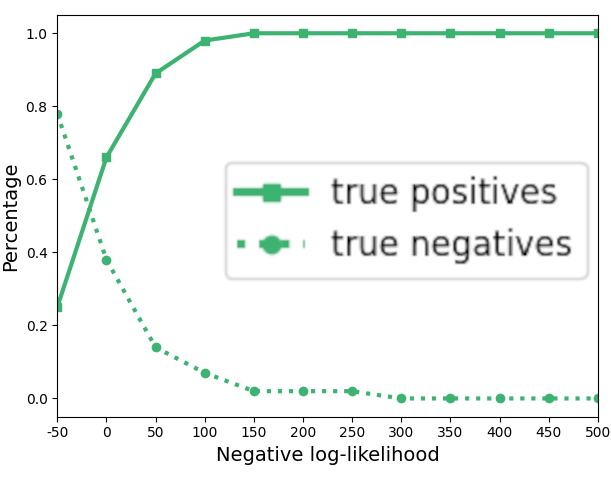}
  \caption{GMM torso rotation}
  \label{fig:sub-tsne-time}
\end{subfigure}
\begin{subfigure}{.4\columnwidth}
  \includegraphics[width=\columnwidth]{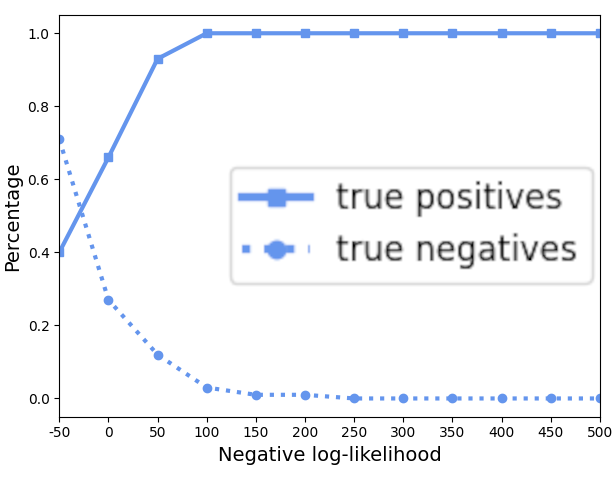}
  \caption{GMM flank stretch}
  \label{fig:sub-tsne-basic}
\end{subfigure}
\begin{subfigure}{.4\columnwidth}
  \includegraphics[width=\columnwidth]{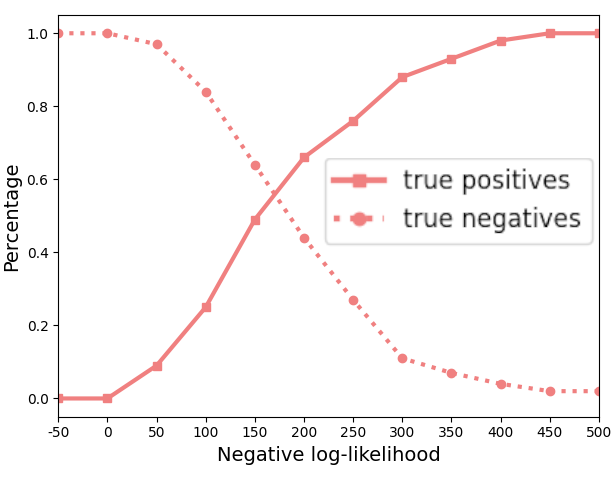}
  \caption{GMM hiding face}
  \label{fig:sub-tsne-mt}
\end{subfigure}
\vspace{-0.5em}
\caption{GMM baseline : true positives and true negatives evolution with respect to different thresholds employed to evaluate a movement as correct or incorrect.}
\vspace{-1em}
\end{figure*}

\subsection{Rehabilitation motion error classification}
For the challenge of error classification, we aim to compare the two baselines for supervised classification. The goal is to classify an observed exercise among 4 classes corresponding to \textit{correct}, \textit{error1}, \textit{error2} and \textit{error3}, for each exercise separately. 
We evaluate the two baselines: 1) GMM-based features combined with a SVM classifier and 2) LSTM classifier. These two approaches are compared in term of classification accuracy. Classification results of GMM-based features combined with SVM classifier is reported in Table\ref{tab:Classification}. For the LSTM classifier, as it may depend on initialization, we run the experiment $10$ times and report mean accuracy with standard deviation. The best accuracy among the 10 runs is also reported.

\begin{table*}[h!]
\caption{Accuracies of GMM-based SVM classifier and the LSTM classifier.}
\label{tab:Classification}       
\begin{tabular}{|l |l |ll|}
\hline
 & GMM \& SVM & \multicolumn{2}{c}{LSTM} |\\
 \hline
Exercise & Accuracy &Accuracy (mean$\pm$std) & Best Accur.  \\
\hline
torso rotation & 27.78 \% & 53.89$\pm$4.82 \% & 64.44 \% \\
flank stretch & 25.32 \% & 31.64$\pm$6.48 \% & 43.03\% \\
hiding face & 33.33 \% & 49.1$\pm$3.28 \% & 56.19\% \\
\hline
\end{tabular}
\end{table*}

 The LSTM classifier obtains much better accuracy for the three exercises. However, the latter approach obtains a maximum accuracy of $64.44\%$ for the torso rotation exercise, showing that recognizing errors is still quite challenging.

\begin{figure*}[h!]
\begin{subfigure}{.36\textwidth}
  \includegraphics[height=\fheight]{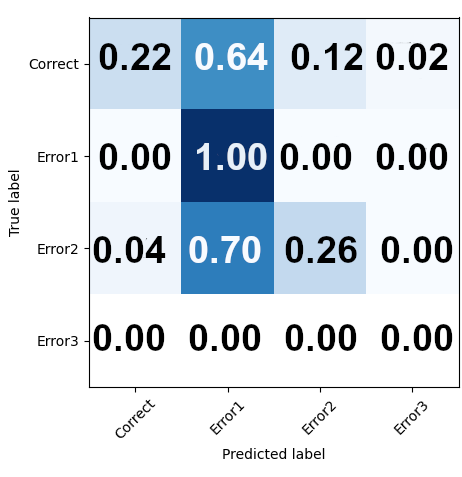}
\vspace{-0.5em}
  \caption{SVM torso rotation}
  \label{fig:sub-class-1}
\end{subfigure}
\begin{subfigure}{.36\textwidth}
  \includegraphics[height=\fheight]{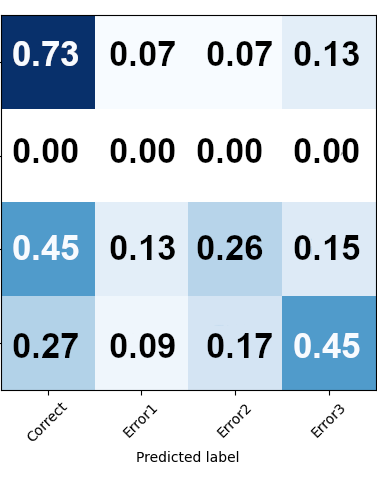}
\vspace{-0.5em}
  \caption{SVM flank stretch}
  \label{fig:sub-class-2}
\end{subfigure}
\begin{subfigure}{.36\textwidth}
  \includegraphics[height=\fheight]{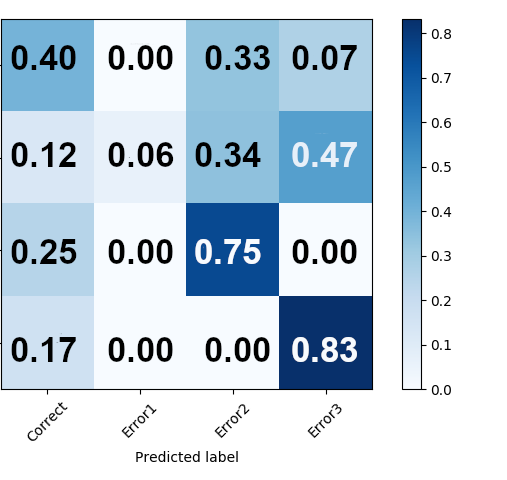}
\vspace{-0.5em}
  \caption{SVM hiding face}
  \label{fig:sub-class-3}
 \end{subfigure}
\vspace{-1.5em}
\caption{Confusion matrices of rehabilitation movements classification using GMM-based features with SVM. Rows full of $0.00$ mean that the corresponding error is not present in test set.}
\vspace{-1em}
\end{figure*}

\begin{figure*}[h!]
\begin{subfigure}[t]{.36\textwidth}
  \includegraphics[height=\fheight]{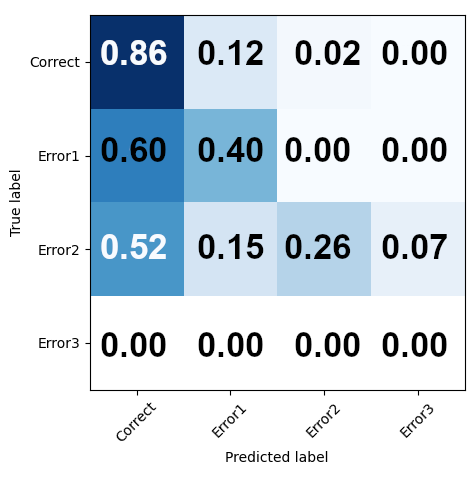}
\vspace{-0.5em}
  \caption{LSTM torso rotation}
  \label{fig:sub-class-1}
\end{subfigure}
\begin{subfigure}[t]{.30\textwidth}
  \includegraphics[height=\fheight]{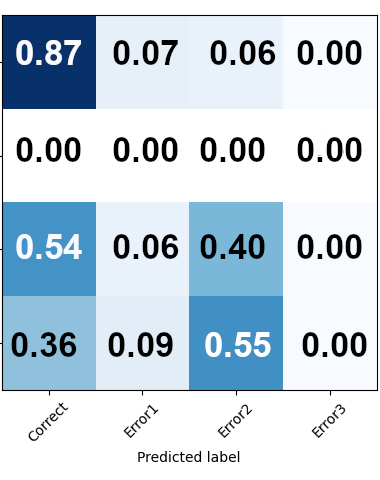}
\vspace{-0.5em}
  \caption{LSTM flank stretch}
  \label{fig:sub-class-2}
\end{subfigure}
\begin{subfigure}[t]{.31\textwidth}
  \includegraphics[height=\fheight]{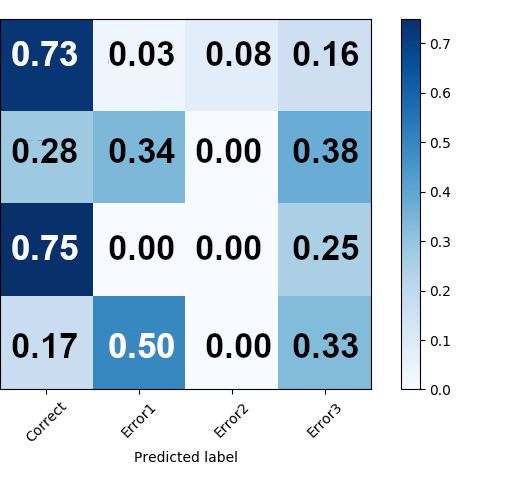}
\vspace{-0.5em}
  \caption{LSTM hiding face}
  \label{fig:sub-class-3}
\end{subfigure}
\vspace{-0.5em}
\caption{Confusion matrices for rehabilitation movements classification using LSTM. Rows full of $0.00$ mean that the corresponding error is not present in test set.}
\label{fig:confmat-lstm-classification}
\label{fig:confmat-svm-classification}
\vspace{-1.5em}
\end{figure*}

From the confusion matrices in Fig. \ref{fig:confmat-svm-classification}, we indeed notice that the majority of correct sequences are not recognized using the GMM-based features with SVM classifier in comparison to the LSTM classifier which is more accurate to recognize correct sequences. Nevertheless, the LSTM classifier is not able to efficiently recognize performed errors for all exercises. This especially explains the lower accuracy obtained for the flank stretch exercise in contrast to the two other exercises, as reported in Table\ref{tab:Classification}. Indeed, for this exercise, fewer test sequences are annotated as correct by the medical expert. Moreover, for the two other exercises (torso rotation and hiding face), we observe a lot of confusion for the error2. As described in  Fig.\ref{fig:errEx}, this error is related to the insufficient medial rotation (yaw) of arms. While it shows that the proposed baseline is not able to handle this medial rotation, it can also be explained by the precision of the skeleton data provided by Kinect.

\vspace{-0.3cm}
\section{Conclusion}
In this paper, we formalized the challenges of automatic coaching of physical rehabilitation exercises from human pose movements as four folds, including motion assessment, error classification, spatial and temporal localization. To benchmark the performances of machine learning algorithms with respect to these four challenges, we introduced the Keraal dataset for human body movement analysis in the context of low-back pain physical rehabilitation. This dataset has been acquired during a clinical study where patients were performing rehabilitation exercises. 
 The limitations of this article are : the dataset includes a small number of exercises and patients, and the baseline algorithms have only standard performance. However, the dataset has detailed annotations. It includes 3D skeleton sequences captured by a Kinect, color video sequences and 2D skeleton data estimated from videos. Moreover, medical expert annotations are associated to each patient's performance for assessment of correctness, recognition of errors, spatio-temporal localization of errors. 
The performance of the two proposed baselines show that the dataset is challenging enough to be a benchmark. We believe it can serve the research community of various fields from computer vision, machine learning, robotics, virtual agents, physical medicine and bio-mechanics, in the long run, allowing the patients to have limited access to perform rehabilitation movements on a regular basis. 

In addition, to further facilitate its use, we evaluate and compare two baseline motion analysis algorithms, pertaining to two different approaches, for the tasks of rehabilitation motion assessment and error recognition. While the experiment introduces how to use the dataset, it also demonstrates that the targeted tasks are still challenging. This suggests that specific and more accurate methods should be designed so as to deeply assess rehabilitation movements and differentiate slight errors from correct sequences. 
This latter investigation is part of our future work. Moreover, we also aim to investigate the challenges of spatial and temporal localization of errors along a motion sequence. Finally, we also want to extend the number of rehabilitation exercises considered by our dataset, as well as annotating more samples.

This dataset allows the development of better intelligent tutoring systems for physical rehabilitation and for physical exercises in general. Its social impact is to enable telemedecine and allow access to better exercising for those with difficult access to rehabilitation centers.

\section*{Acknowledgement}
This research was made possible by the support of the Hi! PARIS Engineering Team. 
\bibliographystyle{IEEEtran}
\bibliography{anrbiblio}

\twocolumn[\section*{Appendix : Articulated figures}]

\begin{figure}
\begin{subfigure}{.24\textwidth}
  \includegraphics[height=0.8\fheight]{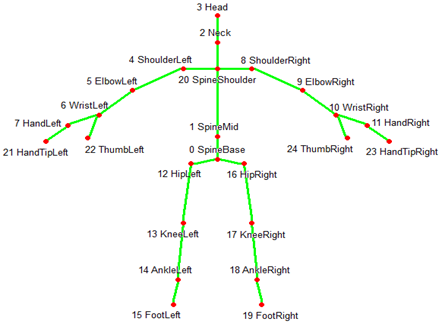}
  \caption{Kinect}
  \label{fig:skeletonKinect}
\end{subfigure}
\hfill
\begin{subfigure}{.24\textwidth}
  \includegraphics[height=0.8\fheight]{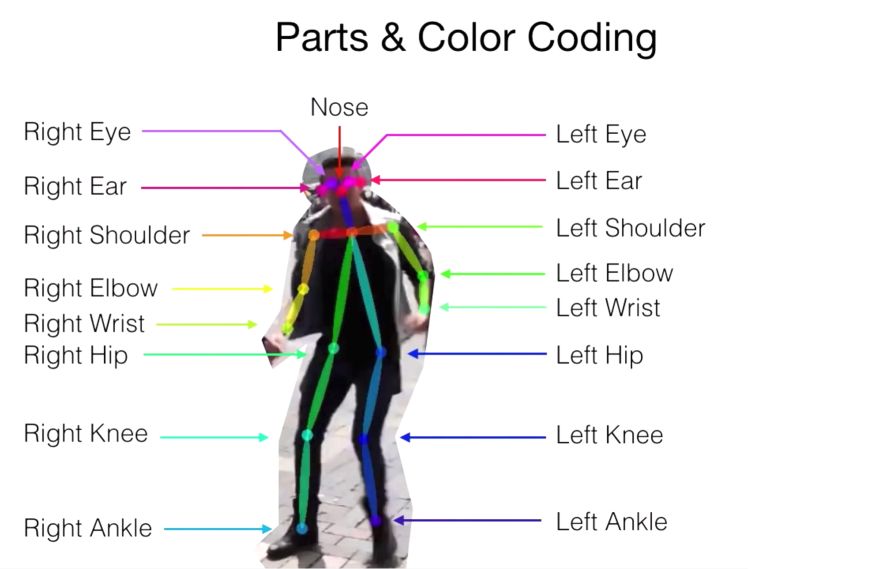}
  \caption{OpenPose Coco \protect\footnotemark
  }
  \label{fig:skeletonOpenpose}
\end{subfigure}
\hfill
\begin{subfigure}{.24\textwidth}
\centering
  \includegraphics[height=0.8\fheight]{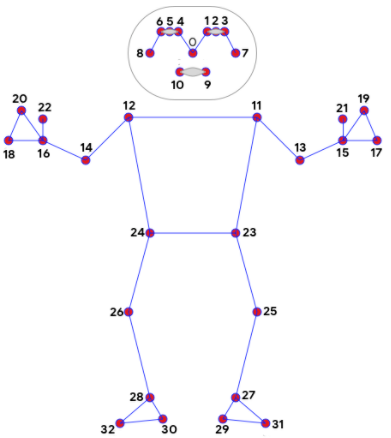}
  \caption{ {33 pose landmarks for BlazePose}}
  \label{fig:skeletonBlazepose}
\end{subfigure}
\hfill
\begin{subfigure}{.24\textwidth}
\centering
  \includegraphics[height=0.8\fheight]{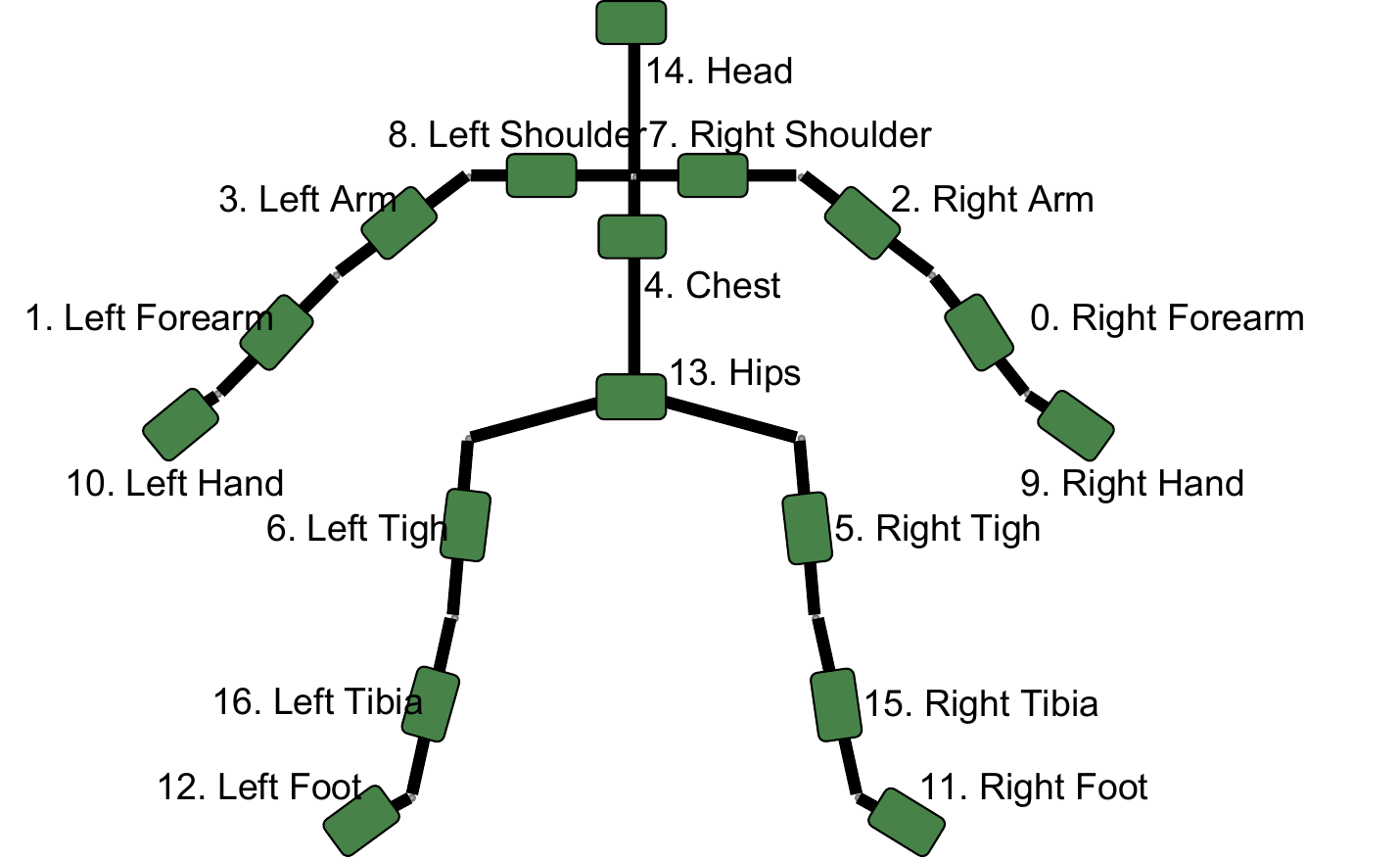}
  \caption{ Vicon}
  \label{fig:skeletonVicon}
\end{subfigure}
\caption{Pose (Skeleton) output formats of the Kinect, OpenPose , {BlazePose} and Vicon.}
\label{fig:skeletons}
\end{figure}

\end{document}